# Scaling Up Biomedical Vision-Language Models: Fine-Tuning, Instruction Tuning, and Multi-Modal Learning

Authors:
Cheng Peng, PhD[1, *]
Kai Zhang, MS[4, *]
Mengxian Lyu, MS[1]
Hongfang Liu, PhD[3]
Lichao Sun, PhD[4, †]
Yonghui Wu, PhD[1,2, †]

Affiliation of the authors:
[1]Department of Health Outcomes and Biomedical Informatics, College of Medicine, University of Florida, Gainesville, Florida, USA

[2]Preston A. Wells, Jr. Center for Brain Tumor Therapy, Lillian S. Wells Department of Neurosurgery, University of Florida, Gainesville, Florida, USA

[3]McWilliams School of Biomedical Informatics, UTHealth, Houston, TX, USA

[4]Department of Computer Science and Engineering, Lehigh University, Bethlehem, PA, USA

[*] Equal contribution

[†] Co-corresponding author

Corresponding author:
Yonghui Wu, PhD
1889 Museum Road, 7th Floor
Gainesville, FL, USA, 32611
Phone: 352-294-8436
Email: yonghui.wu@ufl.edu

Lichao Sun, PhD
113 Research Drive, Building C, Room 325
Bethlehem, PA, USA, 18015
Phone: 610-758-4079
Email: lis221@lehigh.edu

# ABSTRACT

**Objective**

To advance biomedical vison-language model capabilities through scaling up, fine-tuning, and instruction tuning, develop vision-language models with improved performance in handling long text, explore strategies to efficiently adopt vision language models for diverse multi-modal biomedical tasks, and examine the zero-shot learning performance.

**Methods**

We developed two biomedical vision language models, BiomedGPT-Large and BiomedGPT-XLarge, based on an encoder-decoder-based transformer architecture. We fine-tuned the two models on 23 benchmark datasets from 6 multi-modal biomedical tasks including one image-only task (image classification), three language-only tasks (text understanding, text summarization and question answering), and two vision-language tasks (visual question answering and image captioning). We compared the developed scaled models with our previous BiomedGPT-Base model and existing prestigious models reported in the literature. We instruction-tuned the two models using a large-scale multi-modal biomedical instruction-tuning dataset and assessed the zero-shot learning performance and alignment accuracy.

**Results and Conclusion**

The experimental results show that the new models developed in this study outperform our previous BiomedGPT-Base model on 17 of 23 benchmark datasets and achieve state-of-the-art performance on 15 of 23 datasets when compared to previous models reported in the literature. The new models also demonstrated improved ability in handling long text, particularly on text summarization on MIMIC-III dataset and text understanding on SEER dataset, with a remarkable improvement of 4.6~11.4%. Instruction tuning on the scaled models resulted in significant enhancements in zero-shot learning ability and alignment accuracy in following complex instructions across multiple tasks, including image classification, visual question answering, and image captioning. This study develops two vision-language models in the biomedical domain and examines technologies to improve long text content in vision language models through scaling, fine-tuning, and instruction tuning. This study demonstrates the potential of vision language models to integrate multiple data modalities to solve diverse multi-modal tasks in the biomedical domain.

# INTRODUCTION

Artificial intelligence (AI) is rapidly transforming biomedical discovery and healthcare with unprecedented opportunities to enhance disease screening and diagnoses, enable personalized treatments, and speed up new treatment development[1–3]. The increasing heterogeneity and volume of biomedical data, such as genomic, medical images, electronic health records, and patient-generated data, require advanced analytical tools to leverage multiple data modalities[4]. Most recently, transformer-based models have been widely developed to speed up biomedical research and streamline healthcare workflows [5,6]. Large language models (LLMs) such as GPT-4[7], LlaMA[8], and Mistral[9] have been widely used to speed up new treatment development[10] and reduce the burden of healthcare providers through generative tasks such as clinical note summarization[11] and question answering[12]. Vision transformer (ViT) models have shown great potential in multi-modality medical data analysis to automate the detection of pathologies and abnormalities and generate radiology reports from radiology and pathology images such as X-ray, MRI, and CT scans[13–15].

Nevertheless, most early AI models are unimodal single-task systems based on a single data modality, either unstructured clinical notes or medical images [16]. These single-modality models cannot fully leverage complete patient information locked in multi-modal medical data, such as combining imaging data with patient reports. Another critical challenge is the task-specific fine-tuning to adopt models for specific medical applications. Models fine-tuned for a specific task often have limited ability to be adapted to new tasks or different data sources [17,18]. In addition, most models are fine-tuned for classification-based applications with a pre-defined set of labels, diminishing the model's ability for open-ended tasks – a more general form of AI known as artificial general intelligence (AGI). This is particularly important for healthcare applications as physicians need to be in the loop to interact with AI systems to ensure the safety and better outcomes. Therefore, flexible multimodal AI models are needed in the medical domain to fully utilize the modalities of medical data and bring physicians into the loop.

There has been an increasing interest in developing foundation models in the biomedical domain[19,20]. Foundation models are pre-trained on large-scale multitask datasets using self-supervised or unsupervised objectives. Most recently, multimodal vision-language models (VLMs) have emerged[21,22], building on the success of foundational LLMs primarily designed for medical text data. Most successful LLMs support such multimodal applications, such as OpenAI's GPT-4V[23] and Google's Gemini[24], by jointly processing and handling both visual and textual information from different data modalities. Multi-modal VLMs are well-suited for biomedical applications where visual data, such as radiology images, are often accompanied by rich textual data, such as radiology reports. In the medical domain, many multimodal VLMs, such as Med-PaLM M[4], MedFlamingo[25], and LLaVA-Med[26] have been developed to interpret multimodal medical data with reasoning capabilities to advance various biomedical tasks.

We previously introduced BiomedGPT[27], an open-source, lightweight vision-language foundation model designed for diverse biomedical tasks. BiomedGPT employs an encoder-decoder-based transformer architecture to process both images and text, providing a unified framework for various downstream tasks such as medical image classification, visual question answering, medical report generation, and clinical note summarization. BiomedGPT is pre-trained using diverse biomedical datasets, which can be effectively adapted to specific tasks using supervised fine-tuning. While BiomedGPT demonstrated competitive performance across a wide range of biomedical visual and textual tasks, our previous experimental results also showed limitations. First, the model size of the original BiomedGPT is relatively small (up to 182 million parameters) compared with existing models in the general domain. It is well-known that the performance of LLMs can be improved by scaling up the model size, known as the scale-up law[28,29]. Second, the original BiomedGPT model used primarily short snippets of text from biomedical literature, which may limit the model's understanding of long-range text and the ability to generate coherent long-form text, particularly in summarization and report generation tasks. Third, the text-comprehension capabilities, especially in comparison with those of GPT-4V, are not fully examined. The zero-shot in-context learning and complex text understanding applications are not examined.

This study revisited our BiomedGPT model to further examine the scaling up, fine-tuning, and instruction learning for diverse medical applications. Specifically, (1) we develop two new BiomedGPT models using a larger dataset and scale up the model size from 182 million parameters to BiomedGPT-Large with 472 million parameters and BiomedGPT-XLarge with 930 million parameters; (2) we perform a comprehensive evaluation of the three models using comprehensive biomedical downstream tasks, including image classification, text understanding, text summarization, text question answering, vision question answering (VQA), and image captioning; (3) we develop instruction-tuned BiomedGPT models using a new, large-scale, high-quality, multimodal biomedical instruction tuning dataset, and thoroughly assess the zero-shot and transfer learning abilities, and alignment accuracy on three multi-modal tasks, including prompt-based image classification, VQA, and image captioning. This study provides valuable insights into the scale-up of multimodal VLMs as well as multitask instruction tuning of VLMs and provides a comprehensive benchmark for evaluating biomedical VLMs.

## METHODS

Based on the original training data of BiomedGPT, we increased the training data size and replaced the original short-form text with large-scale, long-form text from the PubMed literature. Using this new multimodal dataset, we developed two new models including BiomedGPT-Large and BiomedGPT-XLarge. We conducted fine-tuning and instruction-tuning using the new models and evaluated them on diverse biomedical tasks, including image classification, text understanding, text summarization, QA, VQA, and image captioning. We also assessed the zero-shot capabilities after instruction tuning with an instruction-following dataset. **Figure 1** provides an overview of the study design.

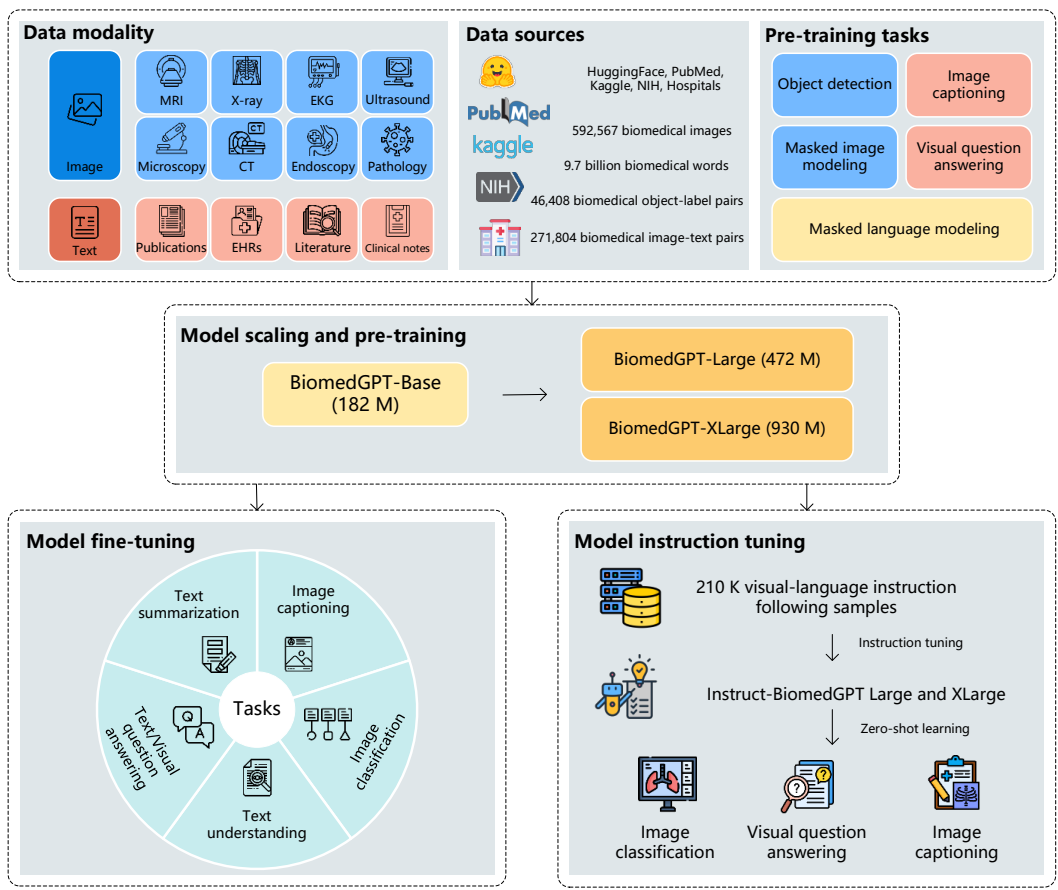

**Figure 1.** An overview of the study design, including model and dataset scaling, model fine-tuning and evaluation on diverse downstream multimodal biomedical tasks, model instruction tuning, and assessment on zero-shot leaning

capabilities. (MRI: Magnetic Resonance Imaging, EKG: Electrocardiogram, CTL: Cytotoxic T Lymphocyte, EHR: Electronic Health Record)

**BiomedGPT architecture**

We implemented BiomedGPT using an encoder-decoder-based transformer architecture, which is widely used to integrate multiple data modalities in a unified model. This allows BiomedGPT to process various data modalities in a unified manner for efficient adaptation across different tasks.

Specifically, the BiomedGPT architecture employs a BERT-style encoder[30] and a GPT-style autoregressive decoder[31]. The encoder is designed to process text inputs, which are tokenized using byte-pair encoding (BPE), to learn contextualized representations through multiple layers of self-attention. For each visual input patch, we obtain its feature vector using the first three blocks of a ResNet[32]. The ResNet module is jointly trained with the transformer module. Meanwhile, a pre-trained VQ-GAN[33] is used to discretize images and provide discrete token sequences as targets during the pretraining phase. Both visual and textual embeddings are augmented with positional embeddings and passed through the shared transformer encoder to construct joint representations of multimodal information. The decoder then processes the information in a causal manner, autoregressively generating the output sequence. All inputs and outputs, including text, image patches, and object bounding boxes, are represented with tokens derived from a unified and finite vocabulary. The unified vocabulary and shared transformer structure are critical for BiomedGPT handling multi-modal inputs and outputs across multiple diverse tasks.

**Scaling up the dataset and model size**

The dataset used in our previous research composes of a set of diverse biomedical data from the following sources: images with diverse modalities such as microscopy, ultrasound, X-ray, CT, MRI, PET, fundus, and dermoscopy, as well as text data from clinical notes, literature abstracts, and object labels. The text data used for pretraining in our previous study was primarily short snippets, with over 99% of the instances having a median length of 19 words and a total number of approximately 9.7 billion words. To enrich the short, small-scale text data, we augmented the pretraining data with large amounts of long-form text data from the subsets of PubMed abstracts and PubMed Central. To ensure the data quality, we conducted a comprehensive data cleaning by removing duplicates and handling missing values. As a result, the text dataset was significantly increased to 14.5 billion words, with a median length of 50.

In our previous work, we developed three versions of BiomedGPT, including BiomedGPT-Small with 33 million parameters, BiomedGPT-Medium with 93 million parameters, and BiomedGPT-Base with 182 million parameters. In this study, we scale up the BiomedGPT model by increasing the number of transformer layers and the hidden dimensions within each layer and develop BiomedGPT-Large with 472 million parameters and BiomedGPT-XLarge with 930 million parameters. The specific configuration and number of parameters of the original and scaled models are summarized in the Supplementary **Table S1**. The scaled models were pre-trained using sequence-to-sequence (seq2seq) learning with a diversified mix of tasks, including masked image modeling (MIM), masked language modeling (MLM), object detection (OD), image captioning, and visual question answering (VQA), and each task is defined with a custom instruction.

**Adopt BiomedGPT through fine-tuning and instruction-tuning**

**Model fine-tuning:** We fine-tuned BiomedGPT models and evaluated performance using a diverse set of multi-modal tasks, for two text-only tasks including text understanding and text summarization, two vision-language tasks including vision question answering (VQA) and image captioning, and a vision-only task - image classification. These tasks were selected to cover a wide spectrum of biomedical applications and data modalities. For fine-tuning, we used a unified sequence-to-sequence approach where the task is formatted using a custom instruction that mirrors those in the pretraining workflow, thereby maintaining consistency and efficiency in model adaptation. We summarized the details of the specific datasets used for each task in **Table 1**. Compared with our previous work, we included two new datasets for question answering and two new datasets for multi-image captioning.

**Table 1**. Modality, tasks, datasets, data source, and instructions used in the BiomedGPT model fine-tuning.

| Modality | Task | Dataset | Data source | Instruction |
|---|---|---|---|---|
| Image-only | Image classification | PathMNIST[34] | Colon Pathology | **[Image]** What does the image describe? |
| | | DermaMNIST[34] | Dermatoscope | |
| | | RetinaMNIST[34] | Fundus Camera | |
| | | PneumoniaMNIST[34] | Chest X-Ray | |
| | | BreastMNIST[34] | Breast Ultrasound | |
| | | BloodMNIST[34] | Blood Cell Microscope | |
| | | OrganCMNIST[34] | Abdominal CT | |
| Language-only | Text understanding | MedNLI[35] (Natural language inference) | Clinical notes | Can text1 "**{Text1}**" imply text2 "**{Text2}**"? |
| | | SEER[36] (Treatment suggestions) | Structured EHR | Please provide treatment suggestion given the patient's information: "**{Text}**". |
| | | TREC'22[37] (Clinical trial matching) | Clinical notes | Please determine the patient's eligibility by comparing the given patient note: "**{Text1}**" and trial details: "**{Text2}**". |

|  |  | MIMIC-III[38] (Mortality prediction) | Clinical notes | What is the predicted outcome for the patient before discharge: "**{Text}**"? |
|  | Text summarization | MeQSum[39] | Natural-language text | What is the summary of the text "**{Text}**"? |
|  |  | HealthCareMagic[40] | Natural-language text |  |
|  |  | MIMIC-CXR[41] | Clinical notes |  |
|  |  | MIMIC-III[38] | Clinical notes |  |
|  | Question answering | PubMedQA[42] | Natural-language text | Your task is to answer biomedical questions using the given context. Only output **yes, no, or maybe** as answer. \n Context: "**{Context}**" Question: "**{Question}**" |
|  |  | MedMCQA[43] | Natural-language text | **{Question}** "**{Options}**" |
| Vision-language | Visual question answering | SLAKE[44] | X-ray, CT scan, MRI, Clinical notes | **[Image]{Question}** |
|  |  | PathVQA[45] | Pathology |  |
|  |  | VQA-RAD[46] | Radiology |  |
|  | Image captioning | IU X-ray[47] | Chest X-Ray | **[Image]** What does the image describe?  (multi-image) **[Image] [Image]** What does the images describe? |
|  |  | MIMIC-CXR[41] | Chest X-Ray |  |
|  |  | PEIR GROSS[48] | Pathology |  |

**Model instruction tuning:** To examine the question understanding and instruction-following capabilities of BiomedGPT models, we further implemented instruction tuning, a supervised learning technique that has shown promise in aligning LLMs with human instructions. We used a large-scale, open-source multimodal biomedical visual instruction tuning dataset, BioMed-VITAL[49], which generates and selects instruction-following data aligned with clinician preference for visual instruction tuning. This dataset has demonstrated higher quality and better training outcomes for the biomedical vision-language model compared with existing open visual instruction tuning datasets, such as LLaVA-Med, used in our previous work. Specifically, the dataset is based on the PMC-15 dataset [50], and the GPT-4-vision-preview API is used to generate multi-round QA instructional data through a two-stage clinician preference alignment process. We used the largest version of the BioMed-VITAL dataset, which contains 210K language-image instruction-following samples. For instruction tuning, we adopted an open-vocabulary setting, diverging from the traditional VQA approach, which uses a predefined answer set during training and inference. This approach allowed BiomedGPT to function without a predefined set of answers.

We evaluated the zero-shot capabilities of the instruction-tuned BiomedGPT models to assess the model's ability to understand and answer biomedical questions in a free-form manner without task-specific fine-tuning. Our previous work mainly focused on the evaluation of VQA tasks using a small subset from the VQA-RAD dataset. In this study, we conducted a more comprehensive evaluation of the zero-shot learning abilities of the instruction-tuned models in comprehensive tasks and datasets, including prompt-based image classification on RSNA pneumonia detection, visual question answering on radiology and pathology images, and image captioning on chest x-ray report generation.

**Experimental Settings**

We represented both images and text as tokens drawn from a unified and finite vocabulary. For visual inputs, we used a frozen image quantization and an object descriptor to discretize images and objects, respectively. The bounding boxes of objects in an image were expressed as sequences of location tokens in the format of integers. Text outputs, including object labels and summarizations, were encoded using BPE tokens. The total vocabulary size consisted of 59,457 tokens, including 50,265 language tokens, 1,000 location tokens, and 8,192 vision tokens. We used VQ-GAN with a patch size of 8 and a vocabulary size of 8,192 for the visual inputs. During pretraining, we randomly subsampled 196 image patches, and the maximum model input length was truncated to 512.

We implemented the models using Python (version 3.7.4) and the PyTorch framework (version 1.8.1, CUDA 12.2) with the sequence-to-sequence toolkit, fairseq (version 1.0.0). We used a distributed data parallel (DDP) strategy to distribute the training workload across multiple GPUs. For model pretraining, fine-tuning, and instruction tuning, we utilized 8 NVIDIA A100 GPUs each with 80 GB of memory. We adopted the same hyperparameters used in our previous BiomedGPT study to train BiomedGPT-Large and BiomedGPT-XLarge. The two models were initialized using the pre-trained weights of OFA-Large and OFA-Huge[51], respectively. For fine-tuning, we adopted a grid search approach to optimize key hyperparameters, including the learning rate, the training batch size, the training epoch, and the warmup ratio. We adopted decoding strategies such as beam search to improve generation quality during inference. For instruction tuning, the pretrained BiomedGPT-Large and BiomedGPT-XLarge were trained for 20 epochs. We compared the configuration of model scaling up in the Supplementary **Table S2**.

**RESULTS**

**Table 2** presents the results of fine-tuned BiomedGPT-Large and BiomedGPT-XLarge across six tasks including image classification, text understanding, text summarization, textual and visual question answering (VQA), and image captioning. We compared our models with existing top systems previously reported in literature, referred to as the baseline, and our previous generation of BiomedGPT-Base model.

**Table 2**. Comparison of fine-tuned BiomedGPT-Large and BiomedGPT-XLarge models with previous reported multimodal VLMs.

| Task | Dataset | Metric | SOTA | | BiomedGPT | | |
|---|---|---|---|---|---|---|---|
| | | | Model | Result | Base | Large | XLarge |
| Image classification | PathMNIST | Accuracy | BiomedCLIP[50] | 0.910 | 0.956 | 0.961 | **0.973** |
| | DermaMNIST | | MedViT-L[52] | 0.723 | 0.866 | 0.872 | **0.882** |
| | RetinaMNIST | | MedViT-L | 0.891 | 0.909 | 0.912 | **0.935** |
| | PneumoniaMNIST | | BiomedCLIP | 0.930 | 0.949 | 0.957 | **0.958** |
| | BreastMNIST | | BiomedCLIP | 0.822 | 0.795 | 0.797 | **0.833** |
| | BloodMNIST | | BiomedCLIP | 0.979 | 0.987 | 0.982 | **0.991** |
| | OrganCMNIST | | BiomedCLIP | 0.925 | 0.910 | 0.921 | **0.935** |
| Text understanding | MedNLI | Accuracy | SciFive[53] | **0.856** | 0.838 | 0.831 | 0.842 |
| | SEER | | BioGPT[54] | 0.459 | 0.500 | 0.511 | **0.535** |
| | TREC'22 | | LLaVA-Med[26] | 0.487 | 0.852 | 0.877 | **0.879** |
| | MIMIC-III | | UMLSBERT[55] | 0.873 | 0.890 | 0.893 | **0.899** |
| Text summarization | MeQSum | ROUGE-L[56] | BioBART-L[57] | 0.532 | 0.523 | 0.505 | **0.556** |
| | HealthCareMagic | ROUGE-L | BART-L[58] | **0.447** | 0.420 | 0.416 | 0.432 |
| | MIMIC-CXR | ROUGE-L | RadAdapt[59] | 0.445 | 0.444 | 0.456 | **0.471** |
| | | F1-Radgraph[60] | | 0.418 | 0.451 | 0.468 | **0.482** |
| | MIMIC-III | ROUGE-L | MedPaLM M (562 B)[4] | 0.320 | 0.307 | 0.322 | **0.342** |
| | | F1-Radgraph | | 0.347 | 0.312 | 0.326 | **0.351** |
| Question answering | PubMedQA | Accuracy | Medprompt[61] | **0.820** | 0.634 | 0.698 | 0.750 |
| | MedMCQA | | Med-PaLM 2 [12] | **0.723** | 0.539 | 0.595 | 0.640 |
| Visual question answering | SLAKE | Accuracy | BiomedCLIP | 0.854 | 0.861 | 0.868 | **0.881** |
| | PathVQA | | CLIP-ViT[62] | **0.636** | 0.581 | 0.618 | 0.624 |
| | VQA-RAD | | MedVInT-TD[63] | **0.816** | 0.732 | 0.769 | 0.794 |
| Image captioning | IU X-ray | CIDEr[64] | PPKED[65] | 0.351 | 0.401 | 0.395 | **0.408** |
| | PEIR GROSS | | CoAttention | 0.329 | 1.227 | 3.225 | **3.421** |
| | MIMIC-CXR | | MedPaLM M[4] | **0.262** | 0.262 | 0.234 | 0.237 |
| | IU X-ray (multi-images) | GREEN[66] | MAIRA-2[67] | **0.474** | 0.369 | 0.391 | 0.463 |
| | | F1-RadGraph | | 0.274 | 0.239 | 0.254 | **0.288** |
| | | F1-Chexbert[68] | | **0.476** | 0.311 | 0.395 | 0.412 |
| | MIMIC-CXR (multi-images) | GREEN | | 0.304 | 0.313 | 0.331 | **0.343** |
| | | F1-RadGraph | | 0.210 | 0.197 | 0.219 | **0.257** |
| | | F1-Chexbert | | 0.515 | 0.409 | 0.448 | **0.520** |

The best scores are highlighted in bold. SOTA: state-of-the-art.

For biomedical image classification, we evaluated our models using the MedMNIST-Raw benchmark datasets. As shown in **Table 2**, we observed a consistent improvement in classification accuracy as model size increased across almost all datasets. Specifically, BiomedGPT-XLarge outperformed the baselines across all seven MedMNIST-Raw datasets, achieving accuracy improvement of 1~15.9%, and BiomedGPT-Large model achieved an accuracy improvement of 0.3~14.9 % compared with the baseline models on five out of seven datasets. and 0.4 ~ 3.8% Compared to our previous BiomedGPT-Base model, BiomedGPT-Large improved accuracies by 0.3~1.1% across all datasets, and BiomedGPT-XLarge further increased by 0.2~3.5%, with the largest gain of 3.5% on BreastMNIST dataset. Overall, the BiomedGPT-XLarge model achieved the best performance across all datasets with various modalities. Notably, BiomedGPT-XLarge outperformed the baseline models with remarkable improments of 15.9%, and 6.3% on dermatology (DermaMNIST dataset) and pathology (PathMNIST dataset) image classification tasks, respectively. For the breast image classification task (BreastMNIST dataset), our previous BiomedGPT-Base model did not match the baseline systems (0.795 versus 0.822), while the scaled BiomedGPT-XLarge model outperformed the baseline with a margin of 1.1%, which is 3.8% of accuracy improvement compared with BiomedGPT-Base.

For the text understanding task, on three of four text understanding datasets, including the SEER, TREC'22, and MIMIC-III, both BiomedGPT-Large and BiomedGPT-XLarge surpassed the baseline models, achieving accuracies improvement of 2~39% and 2.6~39.2%, respectively. Among BiomedGPT models, BiomedGPT-Large outperformed our previous BiomedGPT-Base by 0.3~2.5% on three of four datasets; BiomedGPT-XLarge outperformed BiomedGPT-Large and achieved the best performance on all datasets, with accuracy improvement of 0.4~3.5% and the largest gain of 3.5% on TREC'22 dataset. The 39.2% accuracy improvement over LLaVA-Med on TREC'22 and 7.6% over BioGPT on SEER highlight the model's ability to process complex clinical notes and provide treatment suggestions. For the natural language inference task, our three BiomedGPT models did not outperform the baseline SciFive, with gaps of 2.1%, 2.9%, and 1.6%, respectively. Interestingly, BiomedGPT-Large slightly underperformed BiomedGPT-Base on MedNLI (0.831 *vs.* 0.838), suggesting that scaling up may not fully address dataset-specific challenges.

We evaluated BiomedGPT for text summarization on four benchmark datasets including MeQSum, HealthCareMagic, MIMIC-CXR, and MIMIC-III, using the ROUGE-L metric. We further calucalted F1-Radgraph (measures the factual accuracy of the generated radiology report summaries) score for MIMIC-CXR and MIMIC-III datasets. BiomedGPT-XLarge outperformed the baseline model BioBART-L on the MeQSum dataset, with a ROUGE-L improvement of 2.4%. BiomedGPT-XLarge also achieved better ROUGE-L and F1-Radgraph scores compared with previous models on MIMIC-CXR and MIMIC-III datasets, with improvement of 2.6~6.4% and 0.4~2.2%, respectively. Among BiomedGPT models, we observed consistent performance improvements by scaling up the model size for MIMIC-CXR and MIMIC-III datasets. Specifically, BiomedGPT-Large improved ROUGE-L by 1.2% and 1.5%, and F1-Radgraph by 1.7% and 1.4%, respectively. Compared with BiomedGPT-Base, BiomedGPT-XLarge further improved ROUGE-L by 1.5% and 2%, and F1-Radgraph by 1.4% and 2.5%, respectively. Notably, on the MeQSum dataset, while BiomedGPT-XLarge achieved a score of 0.556 and shows a large improvement over BiomedGPT-Base (0.523), the BiomedGPT-Large model shows a slightly lower score of 0.505. We observed a similar performance saturation on the HealthCareMagic dataset, where BiomedGPT-Large achieved a score of 0.416, which is slightly lower than

BiomedGPT-Base. BiomedGPT-XLarge achieved the best score of 0.432, outperforming the two smaller BiomedGPT models.

We evaluated BiomedGPT models on two types of question-answering tasks, including text-only QA and VQA. For the text-only QA task, we evaluated the models using the PubMedQA and MedMCQA datasets. BiomedGPT models did not outperform the baselines, with a gap of 7~18.6% on the PubMedQA dataset and a gap of 8.3~18.4% on the MedMCQA dataset. Nevertheless, we observed performance increases by scaling up the model size. BiomedGPT-Large outperformed BiomedGPT-Base by of 6.4% and 5.6%, and BiomedGPT-XLarge outperformed the smaller BiomedGPT models with better accuracies of 0.750 and 0.640 on the two datasets, which remarkably outperformed BiomedGPT-Base with improvements of 10.1~11.6%. For VQA, all BiomedGPT models outperformed baseline models on three datasets containing text-image pairs: SLAKE, PathVQA, and VQA-RAD. BiomedGPT-XLarge model outperformed the baseline model BiomedCLIP, with an accuracy improvement of 3.7%. However, BiomedGPT models did not outperform the previously reported best model for the other two datasets. Similarly, within BiomedGPT models, we observed consistent performance improvements by scaling up. Specifically, on the SLAKE dataset, the BiomedGPT-XLarge model achieved the highest accuracy of 0.881, with an improvement of 2% and 1.3% compared with BiomedGPT-Base and BiomedGPT-Large, respectively. Similar findings were observed from the PathVQA and VQA-RAD datasets, where BiomedGPT-XLarge achieved better accuracies of 0.624 and 0.794, respectively, compared to 0.581 and 0.732 of BiomedGPT-Base, and 0.618 and 0.769 of BiomedGPT-Large.

We compared BiomedGPT models with previously reported best models using multiple evaluation metrics, including CIDEr, GREEN, F1-RadGraph, and F1-Chexbert. These metrics are widely used for assessing the similarity and consensus between the generated text and the reference text written by medical experts. BiomedGPT-XLarge outperformed previous models on the IU X-ray and PEIR GROSS datasets, with significant CIDEr score improvement of nearly 10 times on the PEIR GROSS datasets, but with slightly lower performance to MedPaLM M on the MIMIC-CXR dataset. For multi-image captioning, BiomedGPT outperformed MAIRA-2 on the MIMIC-CXR datasets for all metrics, and outperformed MAIRA-2 on the IU X-ray dataset under the F1-RadGraph metric, but trailed on IU X-ray for GREEN (0.463 vs. 0.474) and F1-Chexbert (0.412 vs. 0.476). We observed consistent performance improvements through scaling up for all datasets except for the CIDEr score on the single-image MIMIC-CXR dataset. For example, BiomedGPT-Large achieved almost 3 times of improvement over BiomedGPT-Base (3.225 vs. 1.227 CIDEr score) on the PEIR GROSS dataset and a remarkably improvement (0.395 vs. 0.311, 8.4%) of F1-Chexbert on multi-image IU X-ray. BiomedGPT-XLarge further improved the performance on both single-image (CIDEr by 0.3~19.6%) and multi-image (1.2~7.2%) datasets.

In **Table 3**, we evaluated the zero-shot learning performance of the instruction-tuned BiomedGPT models (Instruct-BiomedGPT-Base, Large, and XLarge) using three multi-modal tasks, including image classification, VQA and image captioning, and compared them against previously reported models including LLaVA-Med and Med-Flamingo. For image classification and VQA, performances were measured using accuracy; for image captioning, performances were measured using GREEN and F1-Chexbert. For image classification, all Instruct-BiomedGPT models remarkably outperformed previous models on both datasets, where Instruct-BiomedGPT-XLarge achieved the best accuracies of 0.630 on RSNA pneumonia detection and 0.732 on PathMNIST (cancer tissue). We observed consistent performance improvements among Instruct-BiomedGPT models through scaling up. For RSNA pneumonia detection and PathMNIST cancer tissue datasets, accuracy increased from 0.364 (Base) to 0.532 (Large) and 0.630 (XLarge), and from 0.729 (Base) to 0.731 (Large) and 0.732 (XLarge), on the two datasets, respectively. For zero-shot visual question answering, instruct-BiomedGPT models also demonstrated strong performance compared with the previously reported models, where Instruct-BiomedGPT-XLarge achieved the best performance on both VQA datasets, with significant improvement of 12.5% and 22.5% in accuracy, respectively. The performance improved 1.7% and 3.1% on the two datasets by scaling up BiomedGPT from Base to Large. However, the accuracy increased 2.1% PathVQA but dropped 1.3% on the VQA-RAD dataset by scaling up from Large to Xlarge. For zero-shot image captioning, Instruct-BiomedGPT models outperformed previous models on both datasets, with a remarkable improvement of 6.9~10% in accuracy and 16.6~21.2% in F1-Chexbert score. We observed consistent improvements on the two datasets by scaling up from Base to Large: 4.1% and 4.8%, and from Large to XLarge: 3.9% and 4.9%.

**Table 3**. Zero-shot learning assessment of instruction-tuned BiomedGPT models in three multi-modal tasks.

| Task | Dataset | Metric | Existing models | | Instruct-BiomedGPT | | |
|---|---|---|---|---|---|---|---|
| | | | LLaVA-Med | Med-Flamingo | Base | Large | XLarge |
| Image classification | RSNA pneumonia detection [69] | Accuracy | 0.177 | 0.181 | 0.364 | 0.532 | **0.630** |
| | PathMNIST (cancer tissue)[34] | | 0.493 | 0.180 | 0.729 | 0.731 | **0.732** |
| Visual question answering | VQA-RAD | Accuracy | 0.381 | 0.335 | 0.488 | 0.519 | **0.506** |
| | PathVQA | | 0.489 | 0.385 | 0.676 | 0.693 | **0.714** |
| Image captioning | IU X-ray | GREEN | 0.058 | 0.027 | 0.047 | 0.088 | **0.127** |
| | | F1-Chexbert | 0.177 | 0.131 | 0.246 | 0.294 | **0.343** |

## DISCUSSION

In this study, we developed two biomedical vision-language models, BiomedGPT-Large and BiomedGPT-XLarge using a multimodality biomedical dataset with 30 billion tokens of biomedical text and 1.53 million medical images, following our previous encoder-decoder-based transformer architecture. We systematically examined the traditional fine-tuning performance using the scaled BiomedGPT models across 6 tasks and 23 benchmark datasets, including one image-only task (image classification), three language-only tasks (text understanding, text summarization, and question answering), and two vision-language tasks (visual question answering and image captioning. The experimental results show that our new models outperform our previous BiomedGPT-Base on 17 of 23 benchmark datasets and achieve state-of-the-art performance on 15 of 23 datasets when compared to previous models reported in the literature. We further examined the zero-shot learning and alignment ability of the scaled BiomedGPT models using three tasks including prompt-based image classification, VQA and image captioning on five datasets. The instruct-BiomedGPT models significantly improved the zero-shot instruction following ability compared to the SOTA models, achieving remarkable improvement of 12.5~45.3%.

We continually pretrained the scaled BiomedGPT models based on the OFA model that is a pretrained foundational model in the general domain. We conducted additional experiments to compare pretraining from scratch versus continuous training from the OFA checkpoint. The comparison results from **Table 4** show that the models initialized with the OFA checkpoint consistently outperformed the models pre-trained from scratch across all tasks. We also found that training from scratch requires significantly more training epochs (approximately 2 times) than continued training from OFA checkpoints. Our finding shows that the general-domain knowledge from the OFA model is helpful for domain-specific vision language model applications.

**Table 4.** Comparison between continuous training from existing checkpoints and training from scratch.

| Task | Dataset | Metric | Pretraining Strategy | BiomedGPT-Large | BiomedGPT-XLarge | Pretraining Time Cost (Epochs) |
|---|---|---|---|---|---|---|
| Image Classification | PathMNIST | Accuracy | Continual training | **0.961** | **0.973** | 50 |
| | | | From scratch | 0.946 | 0.952 | 100 |
| Text Summarization | MIMIC-III | ROUGE-L | Continual training | **0.322** | **0.342** | 50 |
| | | | From scratch | 0.288 | 0.305 | 100 |
| Visual Question Answering | SLAKE | Accuracy | Continual training | **0.868** | **0.881** | 50 |
| | | | From scratch | 0.753 | 0.740 | 100 |

The models developed in this study demonstrate improved performance in handling long text content compared with our previous BiomedGPT models. To further examine the impact of adding extensive amounts of long text to train vision language model, we conducted additional experiments using the MIMIC-III dataset for two tasks including mortality prediction and text summarization, where the medium lengths of the context are 405 and 103 words, respectively. We intentionally truncated the text into a maximum of 50 tokens and compared fine-tuning BiomedGPT models using the truncated text versus the full text. As shown in **Table 5**, BiomedGPT-Large and BiomedGPT-XLarge models are more sensitive to the length of text compared with BiomedGPT-Base. This finding indicates that the scaled models pre-trained using longer text sequences are more effective at leveraging longer contextual information during downstream tasks.

**Table 5.** Impact of long-text pretraining on the MIMI-III dataset

| Model | Text length in pretraining | Text length in fine-tuning | Mortality Prediction Accuracy | Text Summarization ROUGE-L |
|---|---|---|---|---|
| BiomedGPT-Base | 50 tokens | Full length | 0.890 | 0.307 |
| | | Truncated (50 tokens) | 0.845 (↓ 4.5%) | 0.276 (↓ 3.1%) |
| BiomedGPT-Large | Full length | Full length | 0.893 | 0.322 |
| | | Truncated (50 tokens) | 0.714 (↓ 17.9%) | 0.242 (↓ 8%) |
| BiomedGPT-XLarge | | Full length | 0.899 | 0.342 |
| | | Truncated (50 tokens) | 0.719 (↓ 18%) | 0.256 (↓ 8.6%) |

We further evaluate the zero-shot performance of the instruction tuned BiomedGPT models with different sizes using randomly sampled 300 cases from the VQA-RAD dataset. As illustrated in **Figure 2** (a), the largest model, Instruct-BiomedGPT-XLarge, achieved higher alignment accuracy (note that alignment does not necessarily indicate correctness, but rather that it is contextually relevant). **Figure 2** (b) shows detailed category-level accuracy. Compared to the small models, that larger models—particularly those at the Large and XLarge scales—exhibit stronger performance in tasks such as abnormality detection, and disease diagnosis. While vision language models show decent performance for some categories, such as abnormality detection and modality recognition, their zero-shot performance still cannot reach human-level capability. This finding is consistent with recent studies examining the zero-shot learning of LLMs[70].

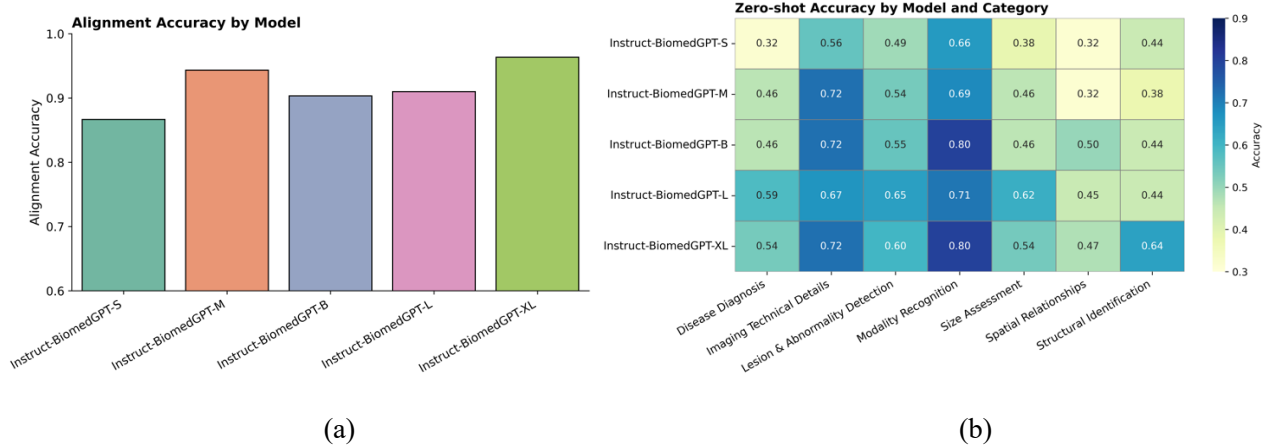

**Figure 2**. Zero-shot performance analysis of instruction-tuned BiomedGPT models. (a) Alignment accuracy across different model scales (Base, Large, XLarge). (b) Zero-shot accuracy breakdown by question category (e.g., Modality Recognition, Lesion & Abnormality Detection, Disease Diagnosis) across model scales, including small, medium, base, large and Xlarge.

This study has limitations. First, the computational cost associated with pretraining and fine-tuning vision language models remains a significant barrier. Although we utilized 8 NVIDIA A100 GPUs, the pretraining of the scaled model still demanded a considerable amount of time, highlighting the practical challenges of scaling up vision language models in resource-limited settings. Second, the images and texts used for curating the instruction-following datasets were primarily taken from the PMC-15M dataset, where the data is not evenly distributed across modalities, with a higher proportion of radiology images compared to gross pathology. This may affect the performance on certain modalities. Third, we mainly used automated machine evaluation scores that are calculated using surface-level strings, which may not reflect the real-world evaluation of humans. Future studies should explore additional data modalities such as video and time-series data.

# CONCLUSION

This study explored the impact of model scaling, fine-tuning, and instruction learning on the multimodal BiomedGPT models, and improved the ability of handling of long texts in vision language models, provides insights into the efficient adoption of vision language models for various biomedical tasks, and demonstrates the potential of vision language models to integrate multiple data modalities in the medical domain.

# ACKNOWLEDGEMENTS

This study was partially supported by grants from Patient-Centered Outcomes Research Institute® (PCORI®) Award (ME-2018C3-14754, ME-2023C3-35934), the PARADIGM program awarded by the Advanced Research Projects Agency for Health (ARPA-H), National Institute on Aging, NIA R56AG069880, National Institute of Allergy and Infectious Diseases, NIAID R01AI172875, National Heart, Lung, and Blood Institute, R01HL169277, National Institute on Drug Abuse, NIDA R01DA050676, R01DA057886, National Cancer Institute, NCI R37CA272473, National Library of Medicine, NLM R01LM011934, and the UF Clinical and Translational Science Institute. The content is solely the responsibility of the authors and does not necessarily represent the official views of the funding institutions. We gratefully acknowledge the support of NVIDIA Corporation and the NVIDIA AI Technology Center (NVAITC) UF program.

## SUPPLEMENTARY MATERIAL

Attached in a separate document.

## DATA AVAILABILITY

All data in this study are publicly available and can be accessed from our previous work. (https://www.nature.com/articles/s41591-024-03185-2)